\documentclass[runningheads]{llncs}
\usepackage[T1]{fontenc}
\usepackage{graphicx}
\usepackage{amssymb}
\usepackage{amsmath}
\usepackage{multirow}
\usepackage{bm}
\usepackage[table]{xcolor}
\usepackage{booktabs}
\usepackage{xcolor} 
\usepackage{url}
\usepackage[colorlinks=true,linkcolor=blue,citecolor=blue,urlcolor=blue,]{hyperref}
\usepackage{marvosym}

\begin{document}
\title{Ultrasound Image-to-Video Synthesis via Latent Dynamic Diffusion Models}
\titlerunning{}
%

\author{Tingxiu Chen\inst{1}\thanks{Equal contribution.} \and
Yilei Shi\inst{1}$^\star$ \and
Zixuan Zheng\inst{1} \and
Bingcong Yan\inst{1} \and
Jingliang Hu\inst{1} \and
Xiao Xiang Zhu\inst{2} \and
Lichao Mou\inst{1}\textsuperscript{(\Letter)}}


%
\authorrunning{T. Chen et al.}
%
\institute{MedAI Technology (Wuxi) Co. Ltd., Wuxi, China\\\email{lichao.mou@medimagingai.com} \and Technical University of Munich, Munich, Germany}
\maketitle            
\begin{abstract}
Ultrasound video classification enables automated diagnosis and has emerged as an important research area. However, publicly available ultrasound video datasets remain scarce, hindering progress in developing effective video classification models. We propose addressing this shortage by synthesizing plausible ultrasound videos from readily available, abundant ultrasound images. To this end, we introduce a latent dynamic diffusion model (LDDM) to efficiently translate static images to dynamic sequences with realistic video characteristics. We demonstrate strong quantitative results and visually appealing synthesized videos on the BUSV benchmark. Notably, training video classification models on combinations of real and LDDM-synthesized videos substantially improves performance over using real data alone, indicating our method successfully emulates dynamics critical for discrimination. Our image-to-video approach provides an effective data augmentation solution to advance ultrasound video analysis. Code is available at \url{https://github.com/MedAITech/U_I2V}.

\end{abstract}
\section{Introduction}
Ultrasound imaging is widely used in clinical practice owing to its non-invasive, radiation-free, and real-time nature. As a primary screening tool, it enables the identification of abnormalities in major organs. However, the accurate interpretation of ultrasound images remains challenging even for experienced practitioners, as healthy and diseased areas in images frequently demonstrate indistinguishable appearances.
\par
Deep learning has emerged as a promising approach for providing suggestions to assist clinicians in their decision making. Ultrasound image classification using deep networks has become a major research area, with state-of-the-art models achieving impressive results~\cite{ult img3,ult img5}. However, since individual ultrasound images provide limited views of lesions, aggregating multidimensional information across an entire ultrasound video is more advantageous for accurate automated diagnosis. Very recently, ultrasound video classification has become a prominent research area~\cite{vid_ult1,vid_ult2,vid_ult3,vid_ult4}. Yet, in contrast to image datasets~\cite{busi,img_data4}, publicly available ultrasound video datasets~\cite{vid_ult2} remain relatively scarce and limited in size, as ultrasound videos are often not recorded and stored. This poses significant challenges in training deep networks for ultrasound video classification.
We consider alleviating this problem by utilizing a large number of existing labeled ultrasound images rather than collecting more ultrasound videos. Synthesizing plausible video sequences from static images is an attractive solution. By doing so, we can augment training data for video classification models and thus improve their performance.
\par
\noindent\textbf{Related work.}
Ultrasound image generation has been explored through physics-based simulators~\cite{img_gen3,img_gen4,img_gen1} and registration-based approaches. Motivated by the success of generative adversarial networks (GANs), recent works have focused on leveraging GANs for ultrasound simulation. For instance, \cite{img_gen5} uses a stacked GAN model for the fast simulation of patho-realistic ultrasound images. \cite{img_gen6} presents a sketch GAN to synthesize editable ultrasound images and introduces a progressive training strategy to generate high-resolution images. However, these methods are limited to enhancing the quantity and diversity of images, not videos.
\par
Video-to-video translation has garnered interest in recent years. For example, \cite{vid_gen4} presents a GAN model to enhance the video quality of handheld ultrasound devices. \cite{vid_gen2} introduces a causal generative model capable of modifying cardiac ultrasound video content based on desired left ventricular ejection fraction (LVEF) values. \cite{vid_gen3} transfers motion from existing pelvic ultrasound videos to static images using key point detection and GANs. For image-to-video synthesis, \cite{vid_gen1} proposes a diffusion model to generate cardiac ultrasound sequences from ultrasound stills. However, this approach requires LVEF as an additional input, thus limiting its application. Moreover, it is computationally expensive (8 NVIDIA A100 GPUs required). In computer vision, \cite{cINN} also focuses on image-to-video synthesis, but on natural images/videos. In this work, we are interested in a computationally efficient ultrasound video synthesis approach operating directly on individual ultrasound images, without requiring additional inputs.

\noindent\textbf{Contributions.} Ultrasound video synthesis from static images is challenging, as images lack dynamic information present in videos. To address this, we propose an approach to compensate between image and video domains for video generation. Our key contributions are as follows:
\begin{itemize}
\item To our knowledge, this work represents the first attempt to leverage image-to-video synthesis to augment training data for ultrasound video classification. We thoroughly investigate the feasibility of this direction.
\item We propose a latent dynamic diffusion model (LDDM) to translate ultrasound images to plausible dynamic sequences in a computationally efficient manner.
\item We demonstrate strong quantitative and qualitative results on the BUSV benchmark. Notably, classification performance improves substantially when training video-based diagnostic models on combinations of real and synthetic data, indicating our generated videos well emulate real dynamics.
\end{itemize}

\section{Method}
Given a starting image $\bm{x}_0$ to serve as the initial condition, our method aims to synthesize a plausible corresponding video clip $\bm{v}$. The inherently under-determined nature of this problem permits the formulation of a multitude of conceivable video predictions based on $\bm{x}_0$. As such, comprehensively modeling video dynamics requires information beyond what is present in the initial seed image. To address this ill-posed problem, we propose to transform a noise $\bm{n}\sim\mathcal{N}(\bm{0},\bm{I})$ into a dynamic representation by means of a conditional diffusion model. This approach learns to harness stochastic noise to compensate for the absence of intrinsic dynamics in the conditioning input image.
\par
\begin{figure}[!t]
  \centering
\includegraphics[width=1\textwidth]{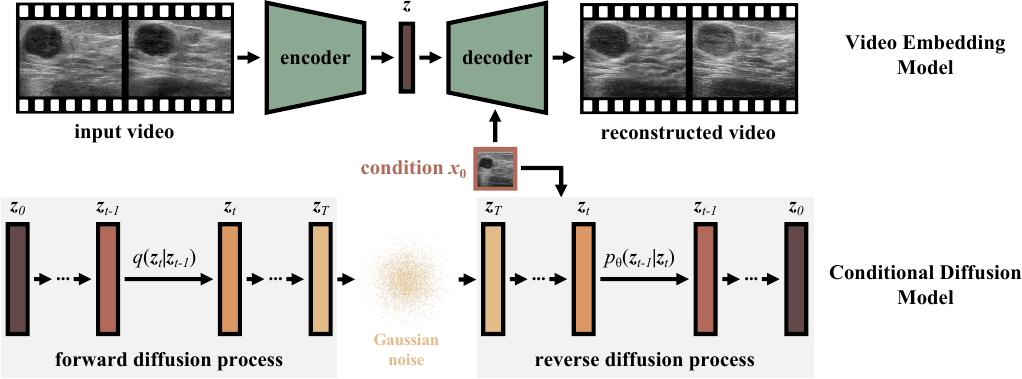}
  \caption{Overview of the proposed LDDM framework.}
  \label{fig:method}
\end{figure}

In this section, we elaborate on the two-stage framework of the proposed LDDM and elucidate the process of video synthesis for enhancing ultrasound video classification. Fig.~\ref{fig:method} shows the overall training pipeline of LDDM. In the first stage, we learn an autoencoding framework to capture intrinsic video dynamics within a low-dimensional latent space. Specifically, an encoder $\mathcal{E}$ compresses an input video into an embedding $\bm{z}$, which is then decoded by $\mathcal{D}$ along with the first frame $\bm{x}_0$ to reconstruct the original clip. In the second stage, we model the latent space encoding dynamics using a conditional diffusion model, where $\bm{x}_0$ provides image context. This allows generating latent embeddings with realistic dynamics from Gaussian noise, conditioned solely on static images at test time. For synthesizing novel videos, given an ultrasound image, we first leverage the trained diffusion model to produce a latent representation $\bm{z}^\prime$. Then, the trained decoder $\mathcal{D}$ transforms the simulated $\bm{z}^\prime$ and the seed image into a synthetic video sequence.

\subsection{Video Embedding}
To acquire encoding and decoding capabilities for videos, we employ an autoencoder framework to represent video dynamics as latent embeddings. Specifically, our encoder $\mathcal{E}$ is implemented as a 3D ResNet~\cite{VAE} (ResNet-18) in order to capture the temporal evolution of dynamics within a video. Additionally, the decoder $\mathcal{D}$ is required to reconstruct the video from the initial frame $\bm{x}_0$ as well as the latent representation of the video. For $\mathcal{D}$, we adopt an architecture consisting of blocks containing AdaIN layers~\cite{karras2019style} to enable the propagation of video information across all scales of the decoder.

\subsection{Conditional Diffusion Model}
Our conditional diffusion model builds on denoising diffusion-based generative modeling~\cite{diffusion1,diffusion2}.
Given a sample from a data distribution $\bm{z}_0\sim q(\bm{z}_0)$, the forward process of our model produces a Markov chain $\bm{z}_1,\ldots,\bm{z}_T$ by progressively adding Gaussian noise based on a variance schedule $\beta_1,\ldots,\beta_T$. At time step $t\in[1,\ldots,T]$, the transition probability is defined as:
\begin{equation}
q(\bm{z}_t|\bm{z}_{t-1})=\mathcal{N}(\bm{z}_t;\sqrt{1-\beta_t}\bm{z}_{t-1},\beta_t\bm{I}) \,,
\end{equation}

where $\beta_t$ is constant. When $T$ is sufficiently large, $\bm{z}_T$ can be well approximated by a standard Gaussian. This suggests that true posterior $q(\bm{z}_{t-1}|\bm{z}_t)$ can be estimated by the learned conditional distribution $p_\theta(\bm{z}_{t-1}|\bm{z}_t)$:
\begin{equation}
p_\theta(\bm{z}_{t-1}|\bm{z}_t)=\mathcal{N}(\bm{z}_{t-1};\mu_\theta(\bm{z}_t),\sigma_t^2\bm{I}) \,,
\end{equation}
where $\sigma_t$ is also constant. 
Sampling then proceeds in this reverse direction, starting from $\bm{z}_T\sim \mathcal{N}(\bm{0},\bm{I})$ and denoising step-by-step via $p_\theta(\bm{z}_{t-1}|\bm{z}_t)$ to obtain $\bm{z}_0\sim p_\theta(\bm{z}_0)$.
\par

To learn $p_\theta(\bm{z}_{t-1}|\bm{z}_t)$, Gaussian noise $\epsilon$ is added to $\bm{z}_0$ to generate samples. Then, conditioned on $\bm{x}_0$, a denoising model $\epsilon_\theta$ is trained to predict $\epsilon$ by minimizing the following mean-squared error loss:
\begin{equation}
\mathcal{L}=\mathbb{E}_{\bm{z},\epsilon\sim\mathcal{N}(\bm{0},\bm{I})}[\|\epsilon-\epsilon_\theta(\bm{z},t,\bm{x}_0)\|^2] \,,
\end{equation}
where $t$ is uniformly sampled from $\{1,\ldots,T\}$.
\par

Given an input video $\bm{v}$, we first obtain its latent dynamic embedding $\bm{z}$ using our trained video encoder $\mathcal{E}$. The size of $\bm{z}$ is $[\Gamma/r, H/r, W/r, C/r]$, where $r$ is a downsampling factor. This is significantly smaller than the original video size, thereby reducing computational requirements. This embedding abstracts away imperceptible details while retaining high-level motion and semantics. We then map $\bm{z}$ to a standard Gaussian variable via the forward diffusion process. The initial frame $\bm{x}_0$ serves as a conditional input to help generate an interpretable dynamic in the latent space.
\par
Since $\bm{z}$ encodes high-level semantic dynamics rather than spatial-temporal details like pixel changes, its dimensions can be much lower than pixel space. As such, the fitting objective of our diffusion model is both meaningful and low-dimensional, easing the generation process. Conditioning on $\bm{x}_0$ allows our model to generate interpretable latent trajectories representing high-level dynamic semantics, as $\bm{x}_0$ provides critical context for plausible video dynamics.

\subsection{Video Synthesis from Single Images}
We synthesize virtual video data using our trained LDDM conditioned on existing images. Specifically, given an image $\bm{x}$, we sample a Gaussian noise and gradually denoise it via the reverse diffusion process to obtain a synthesized latent embedding $\bm{z}^\prime$. We then pass $\bm{z}^\prime$ and $\bm{x}$ through the trained decoder $\mathcal{D}$ to generate a video $\bm{v}^\prime$. Note that the encoder $\mathcal{E}$ is not required at this stage.
\par
Ultrasound videos synthesized in this manner can augment various downstream tasks depending on conditional images. Here, we generate videos from breast ultrasound images to augment video classification models by balancing classes and exposing models to diverse cross-sections. The key advantage is that LDDM learns interpretable high-level dynamics from real videos which manifest in synthesized videos, enabling realistic data augmentation.

\section{Experiments}
\subsection{Datasets}

We utilize the public ultrasound video dataset BUSV~\cite{vid_ult2} to validate the quality of videos generated from initial frames by our LDDM. BUSV comprises 113 malignant videos and 75 benign videos, each with a label indicating the breast lesion type. We create three data splits by randomly selecting 70\%, 50\%, and 30\% of the videos for training, respectively, and use the remaining videos in each split for testing. This allows the comprehensive validation of our approach under different data partition protocols.
\par
Furthermore, we leverage images in the BUSI dataset~\cite{busi} to synthesize more videos for ultrasound video classification. BUSI contains 445 benign lesion images and 210 malignant lesion images. We use all BUSI images to generate video sequences via our trained LDDM.

\subsection{Implement Details}
We implement the proposed LDDM framework using PyTorch and conduct experiments on a single NVIDIA RTX 3090Ti GPU. During training, videos of arbitrary duration are segmented into clips of 48 frames, and we preprocess all frames to have the same dimensions. We then uniformly sample 16 frames from each clip and feed them into the encoder. We utilize the Adam optimizer with a learning rate of $1\times10^{-5}$ for 100 iterations. For the conditional diffusion model, we also adopt the Adam optimizer with a learning rate of 0.01 and a batch size of 16 for 50 iterations. All classification models are trained for 50 epochs until convergence on a single NVIDIA RTX 3090Ti GPU.

\begin{table}[!t]
\centering
\renewcommand\arraystretch{1.0}
\setlength{\tabcolsep}{4pt}
\caption{Quantitative evaluation of video synthesis quality. We compare our LDDM against a previous state-of-the-art approach. Both models are conditioned on the initial frame of a video to generate subsequent frames.}
\begin{tabular}{lcccc}
\midrule[0.85pt]
method     & FVD $\downarrow$ & DTFVD $\downarrow$ & LPIPS $\downarrow$ & DIV $\uparrow$ \\ 
\midrule[0.5pt]
cINN~\cite{cINN} & 58.98 & 0.62   & 0.33   & 2.23 \\ 
LDDM & 53.82 & 0.58   & 0.32   & 1.07 \\ 
\bottomrule[0.85pt]
\end{tabular}
\label{table1}
\end{table}

\subsection{Evaluation Metrics}
We employ four common metrics to evaluate the quality of synthesized videos: Fr\'echet video distance (FVD)~\cite{FVD}, dynamic texture Fr\'echet video distance (DTFVD)~\cite{DTFVD}, learned perceptual image patch similarity (LPIPS)~\cite{LPIPS}, and diversity (DIV)~\cite{DIV} of generated videos. FVD and DTFVD measure holistic similarity between real and synthesized videos, while LPIPS computes patch-level perceptual differences. DIV quantifies variety across generated videos. 
\par
For video classification, we report accuracy and F1 score.

\subsection{Results}
\subsubsection{Video Generation} 
We compare our approach against cINN~\cite{cINN}, a previous state-of-the-art approach that makes use of a normalizing flow model for video generation. As shown in Table~\ref{table1}, LDDM achieves better performance on FVD, DTFVD, and LPIPS. On the DIV metric, cINN produces a higher diversity score. However, we observe that in terms of downstream task performance, cINN underperforms compared to LDDM.

\begin{figure}[!t]
  \centering
\includegraphics[width=1\textwidth]{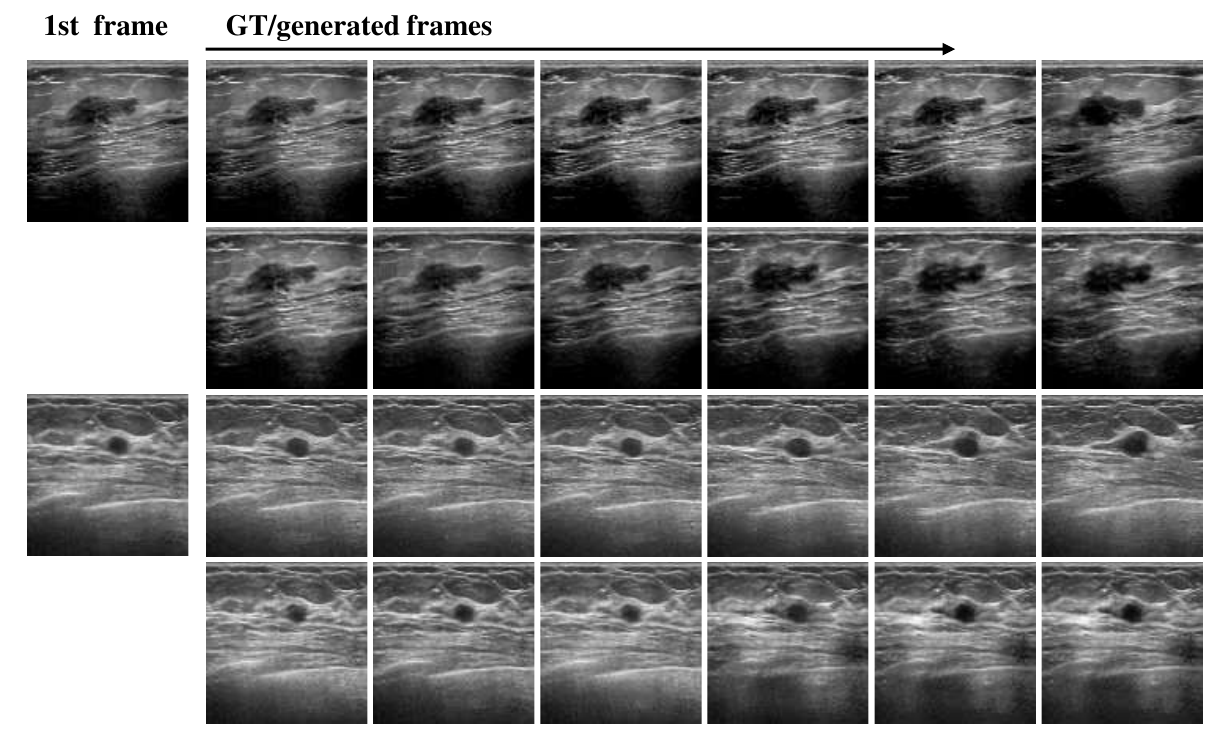}
\caption{Visual examples of ultrasound video generation using the proposed LDDM framework. The bottom row shows synthesized video frames generated by our model conditioned on initial frames from the BUSV dataset, which are unseen during training. The top row depicts ground truth frames for reference. As evident from the visual examples, our LDDM approach is capable of generating realistic ultrasound video sequences that are analogous to real ones.}
  \label{fig:example2}
\end{figure}

\begin{figure}[!t]
  \centering
\includegraphics[width=1\textwidth]{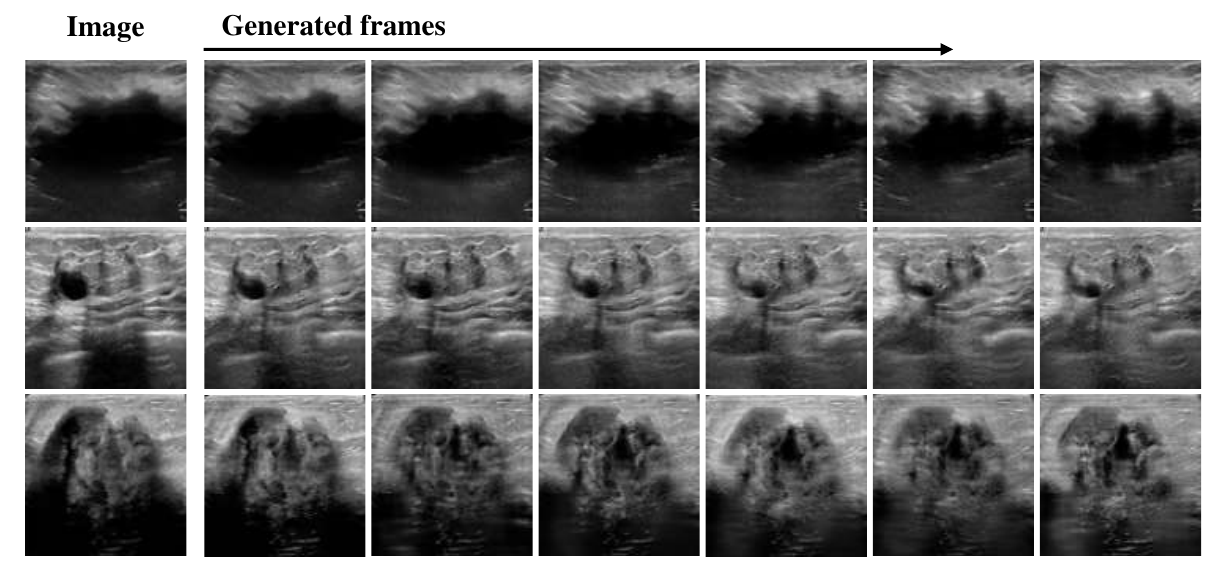}
\caption{Visual examples of generated ultrasound videos based on images from the BUSI dataset.}
  \label{fig:example3}
\end{figure}

This potentially indicates a deviation from realistic motion patterns of nodules. In addition, we show example generation results in Fig.~\ref{fig:example2} and Fig.~\ref{fig:example3}.

\subsubsection{Synthesized Videos for Classification}
Due to the lack of ground truth videos, the quality of generated videos from images in BUSI cannot be directly evaluated using metrics such as FVD. Instead, to demonstrate the interpretability and usability of these synthesized data, we exploit downstream tasks. Specifically, we mix the generated videos with real ultrasound videos to train video classification models. We then evaluate whether the inclusion of these synthetic samples improves models' performance on the BUSV test set.
\par
We use four video classification models: C3D~\cite{C3D}, TSM~\cite{TSM}, TDN~\cite{TDN}, and ATMNet~\cite{ATMNet}. For each model, we compare training with only real videos from BUSV against training with a mixture of real videos and synthesized samples from BUSI. We also provide baselines trained using only synthetic data. Upon thorough analysis in Table~\ref{table2}, we observe significant performance gains when augmenting real data with generated videos. Moreover, as the amount of real training videos decreases, the improvement from adding synthetic samples increases. This indicates that the videos generated by the proposed LDDM are interpretable and beneficial for learning. Our method can thus serve as an effective data augmentation strategy in scenarios with limited access to real medical video data.
\par
These compelling results illustrate that LDDM can produce realistic videos which readily transfer to downstream tasks, ultimately enhancing classification accuracy across various models. The proposed approach shows promise in addressing the scarcity of labeled video data.

\begin{table}[!t]
\centering
\renewcommand\arraystretch{0.9}
\setlength{\tabcolsep}{4pt}
\caption{Evaluation of video classification performance using models trained on different data sources: (1) real ultrasound videos only, (2) synthesized videos conditioned on static BUSI images only, and (3) a combination of real and synthetic data. We compare the downstream task performance of our proposed LDDM approach (black) against cINN (gray).}
\begin{tabular}{lccccccc}
 \multicolumn{2}{c}{}& \multicolumn{2}{c}{real only} & \multicolumn{2}{c}{synthetic only} & \multicolumn{2}{c}{real+synthetic} \\ 
 \midrule[0.85pt]
Method & real data ratio & F1 & Acc & F1 & Acc & F1 & Acc \\
\multirow{6}{*}{C3D} & \multirow{2}{*}{$7\colon3$} & \multirow{2}{*}{70.96} & \multirow{2}{*}{73.21} & \color{gray} 32.89 & \color{gray} 44.64 & \color{gray}74.15 & \textbf{\color{gray} 75.00} \\
 &  &  &  & 59.08 & 58.93 &  \textbf{75.20} & \textbf{75.00} \\
& \multirow{2}{*}{$5\colon5$} & \multirow{2}{*}{64.85} & \multirow{2}{*}{\textbf{68.81}} & \color{gray} 45.42 & \color{gray} 49.46 & \color{gray} 66.67 & \color{gray} 67.74 \\ 
&  &  &   &  62.58 &  64.52 & \textbf{67.92} &  67.74 \\ 
& \multirow{2}{*}{$3\colon7$} & \multirow{2}{*}{60.12} & \multirow{2}{*}{64.61} & \color{gray} 44.47 & \color{gray} 49.23 & \color{gray} 64.56 & \color{gray} 65.38 \\
 &  &  &  &  57.75 &  57.69 &  \textbf{66.08} &  \textbf{66.15} \\ 
\midrule[0.5pt]
\multirow{6}{*}{TSM} & \multirow{2}{*}{$7\colon3$} & \multirow{2}{*}{78.91} & \multirow{2}{*}{78.57} & \color{gray} 67.86 & \color{gray} 67.86 & \color{gray} 76.03 & \color{gray} 76.79 \\
 &  &  &  & 72.83 & 72.83 & \textbf{80.96} & \textbf{80.36} \\ 
& \multirow{2}{*}{$5\colon5$} & \multirow{2}{*}{73.10} & \multirow{2}{*}{73.21} & \color{gray} 66.59 & \color{gray} 66.59 & \color{gray} 76.77 & \color{gray} 76.34 \\
 &  &  &   & 72.30 & 72.30 & \textbf{77.52} & \textbf{77.42} \\ 
& \multirow{2}{*}{$3\colon7$} & \multirow{2}{*}{72.96} & \multirow{2}{*}{73.07} & \color{gray} 57.30 & \color{gray} 57.30 & \color{gray} 71.83 & \color{gray} 72.31 \\
 &  &  &   & 72.59 & 72.59 & \textbf{78.96} & \textbf{79.23} \\
 \midrule[0.5pt]
\multirow{6}{*}{TDN} & \multirow{2}{*}{$7\colon3$} & \multirow{2}{*}{83.03} & \multirow{2}{*}{\textbf{83.93}} & \color{gray} 69.21 & \color{gray} 69.21 & \color{gray} 78.62 & \color{gray} 78.57 \\
 &  &  &  & 71.93 & 71.93 & \textbf{83.66} & \textbf{83.93} \\ 
 & \multirow{2}{*}{$5\colon5$} & \multirow{2}{*}{72.96} & \multirow{2}{*}{72.73} & \color{gray} 69.79 & \color{gray} 69.79 & \color{gray} 76.87 & \color{gray} 77.21 \\
 &  &  &  & 71.93 & 71.93 & \textbf{80.40} & \textbf{78.66} \\ 
 & \multirow{2}{*}{$3\colon7$} & \multirow{2}{*}{63.92} & \multirow{2}{*}{65.38} & \color{gray} 69.07 & \color{gray} 69.07 & \color{gray} 72.15 & \color{gray} 72.67 \\
 &  &  &   & 69.65 & 69.65 & \textbf{75.00} & \textbf{75.38} \\ 
 \midrule[0.5pt]
\multirow{6}{*}{ATMNet} & \multirow{2}{*}{$7\colon3$} & \multirow{2}{*}{72.08} & \multirow{2}{*}{73.21} & \color{gray} 69.52 & \color{gray} 69.52 & \color{gray} 73.34 & \color{gray} 75.00 \\
 &  &  &   & 70.24 & 70.24 & \textbf{78.70} & \textbf{78.57} \\ 
 & \multirow{2}{*}{$5\colon5$} & \multirow{2}{*}{72.70} & \multirow{2}{*}{74.19} & \color{gray} 73.31 & \color{gray} 73.31 & \color{gray} 75.12 & \color{gray} 75.27 \\
 &  &  &   & 77.64 & 77.64 & \textbf{75.62} & \textbf{76.34} \\ 
 & \multirow{2}{*}{$3\colon7$} & \multirow{2}{*}{66.91} & \multirow{2}{*}{66.92} & \color{gray} 56.19 & \color{gray} 56.19 & \color{gray} 66.91 & \color{gray} 66.92 \\
 &  &  &   & 64.97 & 64.97 & \textbf{70.72} & \textbf{72.31} \\ \bottomrule[0.85pt]
\end{tabular}
\label{table2}
\end{table}

\section{Conclusion}
Ultrasound videos provide a more comprehensive and dynamic view of anatomical structures and lesions compared to static ultrasound images. However, acquiring rich ultrasound videos poses significant challenges, as only single images are typically reported and stored in clinical practice. In this work, we propose a generative model called LDDM to synthesize ultrasound videos from ultrasound images using a diffusion model-based approach. By altering the input images, high levels of variation can be introduced when generating such synthetic videos. We demonstrate that improvements in both video generation quality and downstream task performance verify the robustness and practicality of LDDM. An interesting area for future work is prompt-driven image-to-video synthesis, where textual prompts can control video generation.

\begin{credits}
\subsubsection{\discintname}
The authors have no competing interests to declare that are relevant to the content of this paper.
\end{credits}
%
%
%
%

\end{document}